\documentclass[conference]{IEEEtran}

\usepackage{cite}
\usepackage{amsmath,amssymb,amsfonts}
\usepackage{algorithmic}
\usepackage{graphicx}
\usepackage{textcomp}
\usepackage{xcolor}
\def\BibTeX{{\rm B\kern-.05em{\sc i\kern-.025em b}\kern-.08em
    T\kern-.1667em\lower.7ex\hbox{E}\kern-.125emX}}
\begin{document}

\title{Remote Inference of Cognitive Scores in ALS Patients Using a Picture Description}

\author{\IEEEauthorblockN{1\textsuperscript{st} Carla Agurto}
\IEEEauthorblockA{\textit{IBM Research} \\
\textit{Yorktown Heights, NY, USA}\\
carla.agurto@ibm.com}
\and
\IEEEauthorblockN{2\textsuperscript{nd} Guillermo A. Cecchi}
\IEEEauthorblockA{\textit{IBM Research} \\
\textit{Yorktown Heights, NY, USA}\\
gcecchi@us.ibm.com}
\and
\IEEEauthorblockN{3\textsuperscript{rd} Bo Wen}
\IEEEauthorblockA{\textit{IBM Research} \\
\textit{Yorktown Heights, NY, USA}\\
bwen@us.ibm.com}
\and
\IEEEauthorblockN{4\textsuperscript{th} Ernest Fraenkel }
\IEEEauthorblockA{\textit{Massachusetts Institute of Technology} \\
\textit{Cambridge, USA}\\
fraenkel@mit.edu}
\and
\IEEEauthorblockN{5\textsuperscript{th} James Berry }
\IEEEauthorblockA{\textit{MGH Institute of Health Professions} \\
\textit{Boston, USA} \\
jdberry@mgh.harvard.edu}

\and
\IEEEauthorblockN{6\textsuperscript{th} Indu Navar}
\IEEEauthorblockA{\textit{EverythingALS, Peter Cohen Foundation} \\
\textit{Los Altos, USA}\\
indu@everythingals.org}
\and
\IEEEauthorblockN{7\textsuperscript{th} Raquel Norel}
\IEEEauthorblockA{\textit{IBM Research} \\
\textit{Yorktown Heights, NY, USA}\\
rnorel@us.ibm.com}
}

\maketitle

\thispagestyle{plain} 
\pagestyle{plain}

\begin{abstract}
Amyotrophic lateral sclerosis (ALS) is a fatal disease that affects not only movement, speech, and breathing but also cognition. Recent studies have focused on the use of language analysis techniques to detect ALS and infer scales for monitoring functional progression. This paper focused on another important aspect, cognitive impairment, which affects 35-50\%  of the ALS population. In an effort to reach the ALS population, which frequently exhibits mobility limitations, we implemented the digital version of the Edinburgh Cognitive and Behavioral ALS Screen (ECAS) test for the first time. This test, designed to measure cognitive impairment, was remotely performed by 56 participants from the EverythingALS Speech Study\footnote{https://www.everythingals.org/research}. 
As part of the study, participants (ALS and non-ALS) were asked to describe weekly one picture from a pool of many pictures with complex scenes displayed on their computer at home. We analyze the descriptions performed within +/- 60 days from the day the ECAS test was administered and extract different types of linguistic and acoustic features. We input those features into linear regression models to infer 5 ECAS sub-scores and the total score. Speech samples from the picture description are reliable enough to predict the ECAS subs-scores, achieving statistically significant Spearman correlation values between 0.32 and 0.51 for the model's performance using 10-fold cross-validation.

\end{abstract}


\section{Introduction}
Amyotrophic lateral sclerosis (ALS), also known as Lou Gehrig's disease, is a progressive neurodegenerative disease that affects the nerve cells responsible for controlling voluntary muscles. This impairment leads to muscle weakness and eventual loss of movement, speech, and the ability to breathe. ALS does not typically affect senses or bladder and bowel functions; it usually affects adults. The highest incidence is in the age range from 40 to 70. Usually, diagnosis takes about 18 months; on average, life expectancy is 2 to 5 years after diagnosis. Currently, there is no known cure for ALS, and treatment focuses on managing symptoms and improving quality of life. Treatments to slow disease progression have a modest effect. 

To assess functional decline and monitor the rate of progression in ALS patients, clinicians use the ALS functional rating scale - revised (ALSFRS-R), which consists of 12 items that evaluate the functional status of patients in the areas of speech, swallowing, handwriting, cutting food, and handling utensils, dressing and hygiene, turning in bed and adjusting bedclothes, walking, climbing stairs, dyspnea (shortness of breath), orthopnea (difficulty breathing when lying down), and respiratory insufficiency. Each item is scored on a 5-point scale ranging from 0 (unable to perform the task) to 4 (normal performance) \cite{cedarbaum1999alsfrs}. However, this test is not designed to assess cognitive and behavioral changes as many individuals with ALS also experience. Actually, it has been documented that about 35-50\% of ALS patients develop some form of cognitive impairment, with 10-15\% meeting the criteria for diagnosis of frontotemporal dementia (FTD) \cite{phukan2012syndrome,goldstein2013changes,chio2019cognitive,pender2020cognitive}. 

The presence of cognitive impairment complicates the management of ALS symptoms, and survival is shorter for people with both ALS and FTD. Therefore, it is important to identify cognitive and behavioral changes early on and tailor interventions to support individuals with ALS and their families. The Edinburgh Cognitive and Behavioral ALS Screen (ECAS) test \cite{abrahams2014screening} is a screening tool that was developed to assess, in a clinical setting, cognitive and behavioral changes in people with ALS. It assesses five cognitive domains: language, memory, visuospatial ability, executive function, and social cognition, as well as three behavioral domains: apathy, disinhibition, and stereotype.
ECAS is considered a useful tool because it is specifically designed to capture the unique challenges faced by individuals with ALS. The screen includes an assessment of ALS Non-specific functions (recall,  recognition memory, and visuospatial functions) to differentiate cognitive change characteristics of ALS from other disorders common in older adults, such as Alzheimer’s disease (AD). Even though the test is designed to screen ALS patients, it is not limited to only these patients; in fact, to find the normal range for each of the sub-scores, the test was administered to healthy volunteers with no significant neurological or psychiatric history \cite{abrahams2014screening}.

Language analysis, in particular of a free narrative from a participant describing an image, has been widely used to assess cognitive impairment. One of the main reasons is that cognitive impairment presents deficits or changes in language, which are well captured when describing a scene. \cite{cummings2019describing,martinc2021temporal,stegmann2022automated}. This paper also analyzes the picture descriptions from the EverythingALS Speech Study participants. This study recruits ALS participants,  neurologically unaffected controls (related or unrelated to people with ALS), and neurologically unaffected carriers of ALS-causative gene mutations.
Different pictures (one at  a time) with complex stories were shown to the participant in weekly at-home sessions to mitigate the bias caused by learning effects. This will also help to find features that are robust to variance. From the descriptions, we extract different types of linguistic features as well as acoustic features and evaluate the use of these features to infer ECAS total and sub-scores. By automating this evaluation through the use of a picture description, we aim to help the ALS community, which in many situations is overwhelmed with many evaluations and the burden of getting to a clinic.

\section{Methods}

\subsection{Protocol}

EverythingALS is a patient-focused non-profit organization that brings technological innovations and data science to accelerate research toward finding a cure for ALS. The speech study, launched in November 2020, recruits ALS patients and healthy volunteers (to serve as controls). Any adult living in the USA can enroll in this speech study as an ALS or a non-ALS participant. The participants with consistent participation were offered to take the ECAS testing. Every participant documented informed consent prior to participation. By using the modality.ai platform\cite{neumann2021investigating} -a dialog platform for the remote assessment and monitoring of ALS at scale,- the participants performed a series of verbal tasks every week:  sustained vowel phonation,  reading tasks, a measure of diadochokinetic (DDK) rate,  and free speech via a picture description task. The latter task was found to be very useful for assessing cognitive impairment  \cite{cummings2019describing,martinc2021temporal,stegmann2022automated}, and this is the one task we focus on for this study.
Specifically, participants describe a line drawing each week.  Line drawings are rotated to reduce repetition.  Pictures were created following a similar drawing style of the Cookie Theft  \cite{goodglass2001bdae} picture. Participants were also asked to fill out the standard tool for monitoring ALS progression (ALSFRS-R inventory).

A sub-cohort from the EverythingALS Speech Study was selected and asked to take the ECAS test. The ECAS test was administered remotely using a digitized version of the original ECAS test. By being part of the Speech Study, the participants record themselves at home using the same cloud-based platform employed for performing speech tasks \cite{neumann2021investigating}.  
Our inclusion criteria for selecting participants for this study is to have at least one recording of the participants performing the picture description task within +/- 60 days of having administered the ECAS test.

\begin{table}[ht]
    \centering
    \caption{Demographics and clinical variables of the participants (N=56) analyzed in this study}
    \begin{tabular}{cccc}
    \hline
        \textbf{Condition} & \textbf{ALS (N=22)} & \textbf{Non ALS (34)} & \textbf{All (56)} \\ \hline
        \textbf{Gender} & F:10, M:12 & F:25, M:9 & F:35, M:21 \\ 
        \textbf{Age} & 65.3 +/- 6.8 & 62.0 +/- 12.0 & 63.5 +/- 10.3 \\ 
        \textbf{Education years} & 16 +/- 3.1 & 17.6 +/- 2.2 & 17.1 +/- 2.7 \\ 
        \textbf{ALSFRS-R total} & 34.4 +/- 7.0 & 46.4 +/- 2.9 & 41.8 +/- 8.7 \\ 
        \textbf{ALSFRS-R speech} & 3.4 +/- 0.6 & 3.9 +/- 0.3 & 3.8 +/-  0.5 \\ \hline
    \end{tabular}
\end{table}

\subsection{Approach}

\begin{figure} [h]
\centerline{\includegraphics[width=0.5\textwidth]{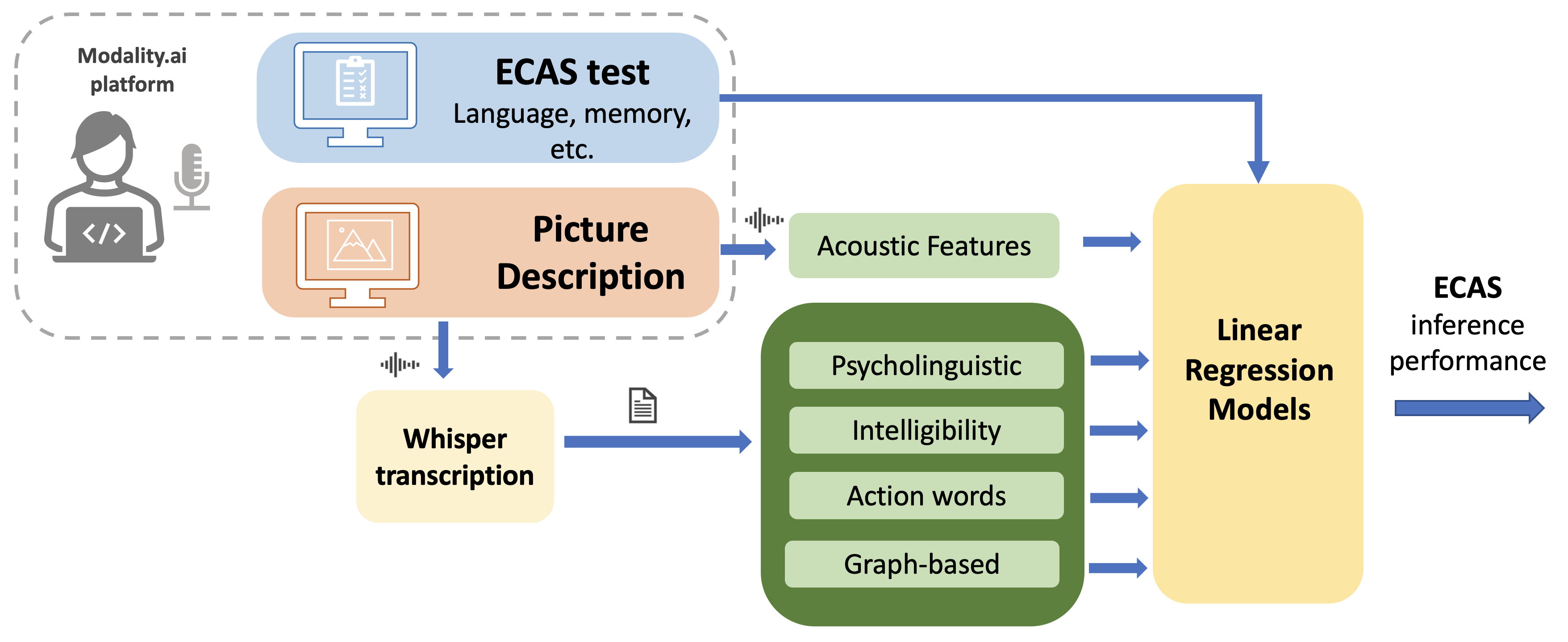}}
\caption{Methodology used to infer ECAS sub-scores and total score. Participants performed the picture description task and ECAS test using modality.ai platform\cite{neumann2021investigating}. Linguistic and acoustic features are extracted and used to train linear regression models to infer the ECAS subs-cores and total score.}
\label{methodology}
\end{figure}

A schematic containing a summary of the approach followed in this paper is shown in Fig. \ref{methodology}. Participants performed the ECAS test and the picture description tasks using the modality.ai platform. Recordings from the participants performing the picture description task were saved in WAV format with a sampling rate of 44.1kHz, and were processed to extract different types of features using the protocol described in \cite{wen2022accelerating}. To perform a thorough characterization of the description, we summarized the acoustic information of the recording using features that focus on prosody, voice quality, noise measurements, and other aspects of the voice using Python \cite{sanner1999python} and Praat\cite{boersma2001praat,boersma2011praat}. A more detailed description of the features used in this paper can be found here \cite{agurto2019analyzing,norel2020speech}.

In addition, we also extracted features of the content of the description of the picture. To do so, the recordings were automatically transcribed using the large language model (LLM) of Whisper Open AI\cite{radford2022robust}. Once we obtained the transcription, we computed 4 different sets of linguistic features. First, we computed psycholinguistic features such as PoS (Part of Speech) using the Python library PoS tagger from NLTK; Then, we calculated lexical richness with Honore's Statistics and Brunet's Index\cite{bucks2000analysis}. Second, since it is well known that ALS speech is affected, we extracted features to measure intelligibility by using as a proxy the difference, measured by metrics in\cite{morris2004and}, in transcription from a small to an LLM \cite{radford2022robust}. The rationale is that both models produce the same or almost the exact transcription for someone with intact speech, whereas the larger model will better understand a person with compromised speech. In addition to these differences, we also extracted the confidence metrics for each language model used in \cite{radford2022robust}. 

The third type of feature is related to action words. We calculated semantic similarity to evaluate the relationship between the descriptions of the pictures by the participants and action or non-action words. Based on the previous literature\cite{hauk2004somatotopic,garcia2014words,garcia2016language,cotelli2018role,norel2020speech}, the following seed words were chosen as action base words for calculating semantic distance: \emph{action, act, move, play,} and \emph{energetic}, or non-action words: \emph{inaction, sleep, rest, sit,} and \emph{wait}. Next, we obtained a numerical representation of all the words using Global Vectors for word representations (GloVe)\cite{pennington2014glove}. Finally, the similarity distance was computed between each word the participant spoke and the seed words. To represent the distribution obtained for each seed word, the following statistical descriptors were calculated from the distances of the participant’s words:  percentiles 5, 50, and 95, interquartile range, skewness, and kurtosis \cite{norel2020speech}.  

Finally, the fourth set is composed of graph features that will measure the lexical complexity of the language used to describe the picture. Different aspects of non-pathological language have been studied using complex network models derived from graph theory\cite{sigman2002global,mota2012speech}. A graph represents a network with nodes connected by edges; in our case, nodes correspond to (unique) words, and edges correspond to adjacent words in the sentence. We then perform basic graph theory characterization of those graphs. Since each picture description has a different number of words, we used windows of fixed size (30, 50, and 100) to compute the graphs and the associated metrics to avoid penalizing longer or shorter descriptions. Part of the metrics computed for each graph is the number of nodes and edges, the mean and standard deviation of the distribution of degrees for each node, the number of loops of sizes 1 to 5, and the largest strongly connected component (LSC). 

Once all the features were extracted, we evaluated the use of each set of features (4 linguistic and 1 acoustic) independently to infer ECAS sub-scores and total score. We trained linear regression models, and performance was measured using Spearman correlation given the non-Gaussian distribution of values in the scores. All models were cross-validated using 10-fold. In addition, we computed a null hypothesis for each model to see if our results were statistically significant.

\section{Results}
\subsection{ECAS scores}
Figure 2a shows the results of the ECAS test in our cohort of 56 participants. We observed that the mean value of ECAS total score in our cohort was 116.3 +/- 8.3, which indicates that overall this was  not a highly impaired cohort (abnormality cut-off value is 105 according to \cite{abrahams2014screening}). We also noticed that the visuospatial score has a low standard deviation, indicating insufficient variance in the samples to create a model to infer this variable. Figure 2b shows the correlation, in our entire cohort (N=56), between the ECAS scores and demographic variables such as age and education years. Overall, positive correlations were observed as expected for education years, especially with the verbal fluency score (r=0.13), and negative correlations were observed for age, especially with the ECAS total score (r=-0.32). In addition to demographics, we also computed the correlations between ECAS scores and  ALSFRS-R speech and total score. Contrary to the correlation values obtained with the demographic variables, we did not observe a clear trend when ECAS scores were compared to the two ALS assessments. For example, ALSFRS-R speech correlates positively with language score (r=0.17) and negatively with memory score (r=-0.28).
We also evaluated the abnormality cut-off values reported in the paper \cite{abrahams2014screening} (using only healthy controls) in our entire cohort. Figure 3 shows all the scores obtained for all the participants divided into two sub-cohorts: ALS (orange dots) and non-ALS participants (blue dots). Cut-off values are depicted using a black horizontal line. Except for language score, for which abnormal cases constitute 45\% for ALS and 35\% for the non-ALS population, abnormal cases in the remaining scores for both sub-cohorts are below 20\%. Specifically, the percentage of impairment of ALS (non-ALS) participants is detailed as follows: verbal fluency 5\% (12\%), which corresponds to a 66 years old participant (4 participants between 68 to 81 years old), executive functions 18\% (0\%), memory functions 18\% (18\%), visuospatial functions 0\% (6\%), and ECAS total score 18\% (12\%).

\begin{figure} [h]
\centerline{\includegraphics[width=0.4\textwidth]{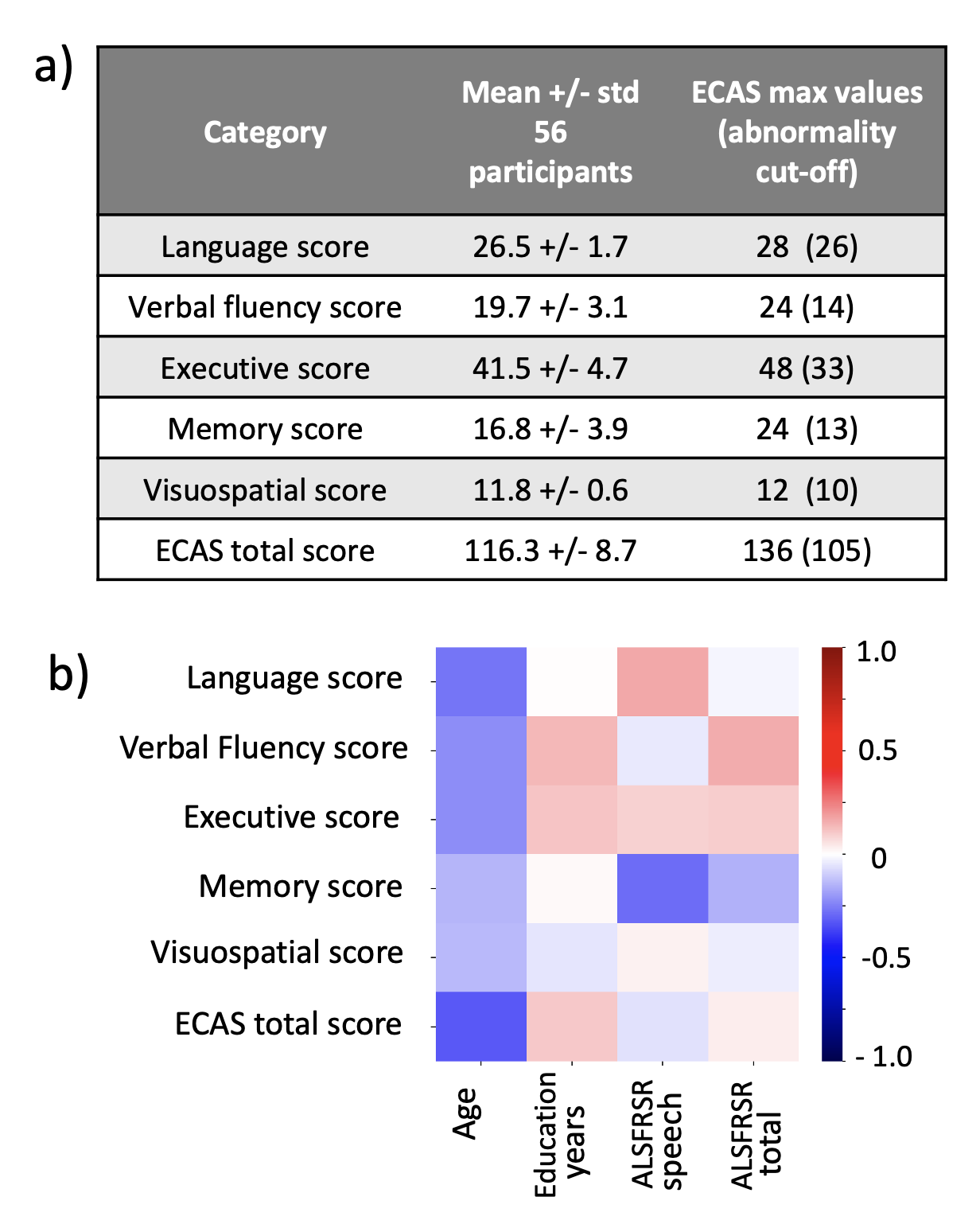}}
\caption{Results of the ECAS test in our cohort of 56 participants. a) Mean and standard deviation values for each of the sub-scores and total score, and maximum and cut-off values for ECAS sub-scores retrieved from \cite{abrahams2014screening}, b) Correlation of the sub-scores with demographic variables (age and education years) and ALSFRS-R speech and total scores.}
\label{ECAS_correlation}
\end{figure}

\begin{figure} [h]
\centerline{\includegraphics[width=0.4\textwidth]{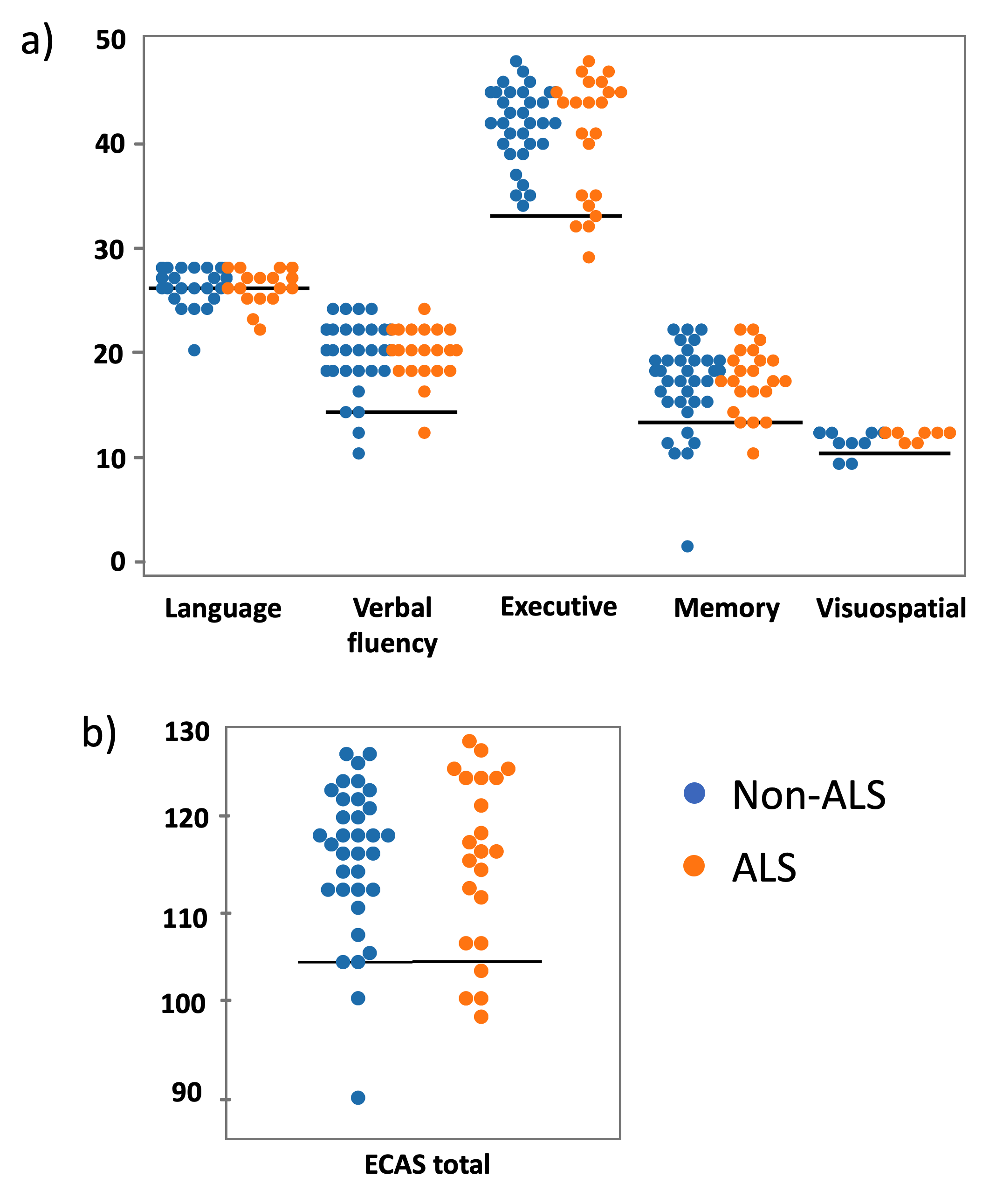}}
\caption{Swarm plots of the results of the (a) ECAS sub-scores and (b) total score in ALS (orange dots) and non-ALS (blue dots) sub-cohorts. Horizontals black lines depict the cut-off values that define abnormality (equal or below the score) in the results\cite{abrahams2014screening}.}
\label{ECAS_swarmplots}
\end{figure}

\subsection{Models' performance}

Table 2 shows the results of the linear regression models for all the pairs of combinations (set of features vs. ECAS scores). In addition, the table shows the p-value in parenthesis to indicate if the result is statistically significant when compared to the null hypothesis. Models with high-performance values were obtained for inferring memory score using acoustic features (r=0.51, p-value=1E-3),  ECAS total score using action words (r=0.45, p-value=1E-3), and language score using graph features (r=0.41, p-value=2E-3). Psycholinguistic and intelligibility features could also infer verbal fluency and executive scores with acceptable correlation values of r=0.34, p-value=4E-3 and r=0.32, p-value=6E-3, respectively. Most feature sets were only valuable to infer one ECAS sub-score except for the action words feature set that can infer the other two sub-scores (memory and verbal fluency) with statistically significant correlation values.
Post-hoc analysis of the relevant features of the models was performed by extracting and plotting the top 5 highest weighted features in the linear regression models (See Fig. 4). Each of the subplots corresponds to the best model of each of the 5 ECAS sub-scores from Table 2. Figure 4 also shows that although we set the limit to only the top 5 features, verbal fluency and ECAS total score required fewer features to infer the score.

\begin{table*}
    \centering
    \caption{Performance results of the implemented linear regression models using Spearman correlation values with p-values in parenthesis.}
    \begin{tabular}{cccccc}\hline
    \hline
        \textbf{} & \textbf{Memory Score} & \textbf{Verbal-Fluency Score} & \textbf{Executive Score} & \textbf{Language Score} & \textbf{ECAS total score} \\ \hline
        \textbf{Acoustic} & 0.51 (1E-3) & 0.1 (2E-1) & 0.08 (3E-1) & 0.19 (8E-2) & 0.12 (1E-1)\\ 
        \textbf{Psycholinguistic} & -- & 0.34 (4E-3) & 0.05 (4E-1) & 0.08 (3E-1) & --\\ 
        \textbf{Intelligibility} & 0.14 (1E-1) & 0.02 (5E-1) & 0.32 (6E-3) & 0.09 (3E-1) & 0.06 (7E-1) \\ 
        \textbf{Graph-based} & 0.12 (2E-1) & 0.19 (8E-2) & -- & 0.41 (2E-3) & 0.24 (3E-2) \\ 
        \textbf{Action words} & 0.39 (1E-3) & 0.26 (2E-2) & -- & -- & 0.45 (1E-3) \\ \hline
     \end{tabular}
     \label{table:performance}
\end{table*}

\begin{figure*}
\centerline{\includegraphics[width=0.9\textwidth]{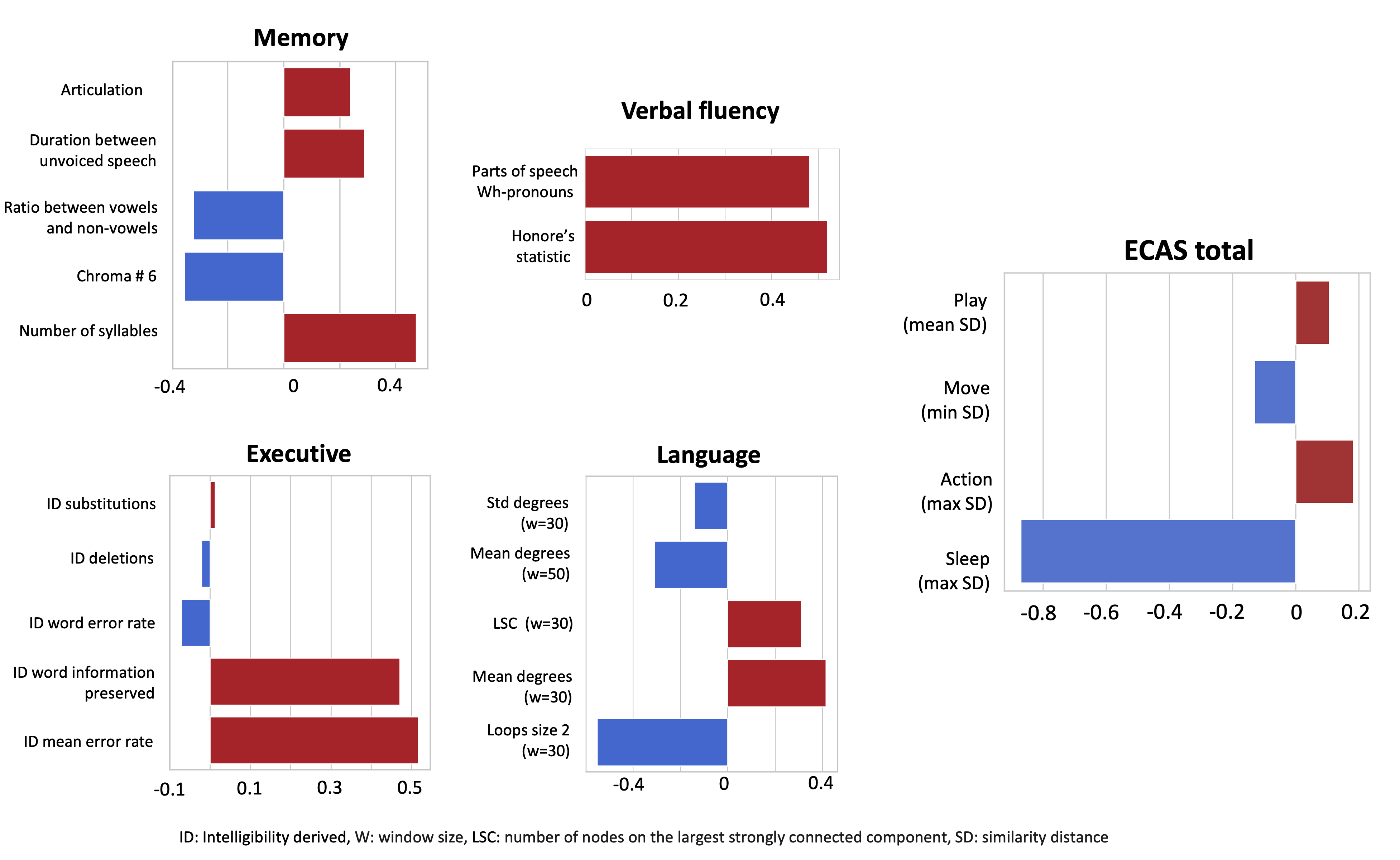}}
\caption{Normalized weights (top 5 features) of the best models used to infer the ECAS sub-scores and total score (see Table 2). A different set of features was used for each model: memory (acoustic features), verbal fluency (psycholinguistics), executive (intelligibility), language (graph-based), and ECAS total (action words). }
\label{weights}
\end{figure*}

\section{Discussion}

We demonstrate that linguistic and acoustic features extracted from a picture description task are informative enough to infer scores that measure cognition (i.e., ECAS) in an ALS and non-ALS population. Our models, which incorporate the description of several images (contrary to other studies), infer all the ECAS sub-scores and total score achieving performance values r\textgreater 0.30. Although we acknowledge that our cohort is small and further validation is required, results are encouraging to continue pursuing the use of automatic methods to monitor cognitive impairment in ALS. Interesting to note that different types of features capture different sub-scores of ECAS, indicating that the ECAS subsections detect distinct impairment types. 

When analyzing the results of the ECAS test in the participants, we observe that ALSFRS-R total score presents a low correlation with the ECAS sub-scores and total score, indicating that the same features used for detecting or monitoring ALS may not be helpful in inferring cognitive impairment. In fact, according to a recent study, the incidence of language impairment for non-demented people with ALS is estimated at 23\%\cite{solca2022prevalence}.
We also observe in Fig. 3 that both sub-cohorts (ALS and non-ALS) have a similar percentage of samples with impairment  based on cut-off values described in \cite{abrahams2014screening}. 
A reason could be that it takes a certain amount of cognitive processing to find, enter, and keep up with an observational study (this is frequently a problem with cognitive evaluation in observational studies).
In addition, we noticed that our non-ALS participants' values differed from those reported for the control group in the ECAS study. A plausible explanation of these results is that in the ECAS study, controls are healthy volunteers that were screened for neurological and psychiatric history, which is not the case for the non-ALS participants from this study. 
 For example, across all the participants, we were surprised to find that one non-ALS participant scored 1 out of 24 for the memory subsection test (See Fig. 3a). This is a highly educated 70 years old female. Other ECAS subsections scores for this participant were found in the normal range. This could be due to delirium due to anxiety, sleep deprivation, medications, illness, or other causes, which can prevent registration of memory, or due to developing dementia;  we will repeat the assessment after 6 months. 
Related to the fact that both cohorts have similar distributions, we also found that the impairment cases are lower than expected for the ALS cohort group \cite{crockford2018specific,chio2019cognitive}. We speculate that the ALS cohort in this study is not progressed enough in the disease. In fact, our cohort only presents 5 participants (23\%) with bulbar onset, a condition that has more incidence of FTD in the literature \cite{gordon2011range,strutt2012cognition,chio2019cognitive}.

To establish a baseline, we analyzed associations between ECAS sub-scores and demographic variables, such as age and education, that are similarly distributed among our sub-cohorts. We found the highest correlation for the pair age and ECAS total score (r=0.32). However, our model based only on action words can achieve a higher correlation of r=0.45. A model incorporating age into the action words model was implemented, achieving (r=0.48, p-value=1E-3), indicating that age information can complement the action words feature set.

Most relevant features were analyzed for each of the best models reported in Table 2. For inferring the memory score, the acoustic feature set was the one that achieved the best performance (see Fig. 4). Among the most relevant features, we observed that articulation and the number of syllables, a proxy for measuring speech tempo, are positively correlated with memory score. Not surprisingly, recall, which is measured in the ECAS memory subsection, is usually associated with speech tempo\cite{riggs1993passage}. On the other hand, the increased ratio between vowels and no vowels is inversely correlated to memory score. We speculate that hesitations or making an effort while recalling words like ``uh" or ``ah" are increasing the number of vowel sounds, thus increasing the ratio.

In the case of relevant features for linguistic-based features, we found that the best model for inferring verbal fluency (measured in the test as listing words beginning with the letter S and listing words beginning with the letter T but with only four letters) uses two psycholinguistic features. Both of them present a positive association with verbal fluency. One of them is Honore's statistics\cite{bucks2000analysis}, which quantifies lexical richness by counting the unique number of words uttered by the speaker (a direct association with verbal fluency), and the other one is the proportion of wh-pronouns (who, what).

Executive score, composed of reverse digit span, alternation, inhibitory sentence completion, and social cognition, is best predicted using the features derived from intelligibility (see Table 2). While most research on executive function and language in neuromotor diseases has been focused on syntactic and word retrieval dysfunction\cite{mckinlay2010execPD, garcia2016language, Abrahams2004WordALS}, the link with phonological dysfluencies is well documented\cite{chan2018phonology}. As it was shown in Fig. 2b, ALSFRS-R speech assessment has a low correlation with the executive score, indicating that it is not just affected speech that is affecting 
this score.

Features derived from the lexical graph analysis are best to infer language score, which is composed of naming, comprehension, and spelling. We observed in Fig. 4 that number of edges per node is a very important feature in the model, in particular for the smaller window size of text (30 and 50). The number of loops of size 2 was the most relevant and negatively correlated with language score. The reason behind this is that this feature can be interpreted as stuttering but with words; for example,  instead of saying \emph{``the boy is looking at the insect"}, they said \emph{``the boy is looking for looking at the insect"}. Finally, we found a positive correlation between LSC and language, which corroborates the findings of other research between LSC and cognitive impairment \cite{mota2012speech, palaniyappan2019speech}

Lastly, the best model for inferring ECAS total score uses mostly 4 features. We observe that action words like ``sleep"(maximum similarity distance) and ``move" (minimum similarity distance) which implies lower movement, are negatively correlated with the total score, while action words like play (median similarity distance) and action (maximum similarity distance) which implies higher movement positively correlated with the total score. The literature supports this finding. Embodied cognition postulates that the motor system influences cognition. In particular, it has been suggested that action words and motor representation of those actions activate the same network in the brain \cite{boulenger2009grasping,peran2010mental,garcia2014words,garcia2016language,norel2020speech,aiello2023rethinking}.

As mentioned before, the limitations of this study include the number of participants and the lack of more impaired cases (samples below the cut-off values). Efforts are underway to recruit more participants from the EverythingALS speech study to take the remote online version of the ECAS. Another limitation of the study is that we are using automatic transcription, which can be prone to errors when transcribing speech from people with dysarthria. Future experiments will also include the use of manual transcriptions to quantify any effect of using automatic transcriptions.

In conclusion, this study provides novel preliminary evidence that a picture description task can be used to implement automatic assessment of cognitive state in the ALS population in a remote setting, a population that, by the nature of the disease, is home-bound at later stages of the disease. We envision that by digitizing health assessment for ALS patients, the life span and quality of life of many of them will improve, given that early detection allows for early management. 

\section*{Acknowledgment}
We are immensely thankful to the individuals who participated in the study and donated their data; this work would not be possible without them.
We thank Michele Merler for running the Large Language Models and Erin Manogaran, Meera Gandhi, and James Sagaser for administering the ECAS test, and Swapnil Harkanth for digitizing the ECAS form for remote administration.

\bibliographystyle{IEEEtran}
\bibliography{ECAS_ALS}
\end{document}